\documentclass{article}

\usepackage{arxiv}

\usepackage[utf8]{inputenc} 
\usepackage[T1]{fontenc}    
\usepackage{hyperref}       
\usepackage{url}            
\usepackage{booktabs}       
\usepackage{amsfonts}       
\usepackage{nicefrac}       
\usepackage{microtype}      
\usepackage{lipsum}

\usepackage{cite}
\usepackage{amsmath,amssymb,amsfonts}
\usepackage{algorithmic}
\usepackage{graphicx}
\usepackage{textcomp}
\usepackage{scalefnt}
\usepackage{tcolorbox}
\usepackage{longtable}
\usepackage{makecell}
\usepackage{verbatim} 
\usepackage{xcolor}
\usepackage{colortbl}

\newcommand{\al}{\textit{et al.~}}
\newcommand{\cons}{SC}
\newcommand{\desi}{SD}
\newcommand{\mode}{SMM}
\newcommand{\test}{ST}
\newcommand{\quali}{SQ}

\definecolor{silver}{RGB}{255,255,255}
\definecolor{lightgray}{RGB}{217,217,217}

\newcommand{\silverrow}{\rowcolor{silver}}
\newcommand{\lightgrayrow}{\rowcolor{lightgray}}

\usepackage{array}
\newcolumntype{L}[1]{>{\raggedright\let\newline\\\arraybackslash\hspace{0pt}}m{#1}}
\newcolumntype{C}[1]{>{\centering\let\newline\\\arraybackslash\hspace{0pt}}m{#1}}
\newcolumntype{R}[1]{>{\raggedleft\let\newline\\\arraybackslash\hspace{0pt}}m{#1}}

\title{Software Engineering for Robotic Systems:a systematic mapping study}

\author{
Marcela G. dos Santos \\
SmArtSE Research Team \\
Université du Québec à Chicoutimi \\
555, boulevard de l’Université, G7H 2B1 \\
Chicoutimi, QC, Canada \\
\texttt{marcela.santos1@uqac.ca} \\
\And

Fabio Petrillo \\
SmArtSE Research Team \\
Université du Québec à Chicoutimi \\
555, boulevard de l’Université, G7H 2B1 \\
Chicoutimi, QC, Canada \\ 
\texttt{fabio@petrillo.com} \\
}

\begin{document}
\maketitle

\begin{abstract}
Robots are being applied in a vast range of fields, leading researchers and practitioners to write tasks more complex than in the past. The robot software complexity increases the difficulty of engineering the robot's software components with quality requirements. Researchers and practitioners have applied software engineering (SE) approaches and robotic domains to address this issue in the last two decades. This study aims to identify, classify and evaluate the current state-of-the-art Software Engineering for Robotic Systems (SERS). We systematically selected and analyzed 50 primary studies extracted from an automated search on Scopus digital library and manual search on the two editions of the RoSE workshop. We present three main contributions. Firstly, we provide an analysis from three following perspectives: demographics of publication, SE areas applied in robotics domains, and RSE findings.  Secondly, we show a catalogue of research studies that apply software engineering techniques in the robotic domain, classified with the SWEBOK guide. 
We have identified 5 of 15 software engineering areas from the SWEBOK guide applied explicitly in robotic domains. The majority of the studies focused on the development phase (design, models and methods and construction). Testing and quality software areas have little coverage in SERS. 
Finally, we identify research opportunities and gaps in software engineering for robotic systems for future studies.
\end{abstract}


\section{Introduction}

Robotic systems (RS) are a complex combination of hardware and software components that interact with the real world. Besides, RS components must be correctly integrated to enable the system to perform as expected. Nowadays, different domains use robots; from industries to healthcare, passed by a critical mission,  robots become more present in our world \cite{AHMAD2016}.   

Recent systematic mappings address software engineering areas in the RS in different SE areas as architecture \cite{AHMAD2016}, requirements engineering \cite{ALBUQUERQUE017}, safety for mobile robotic systems. Besides, there are surveys as \cite{MALAVOLTA2020}, a survey focused on a specific middleware for robotic systems, and \cite{GARCIA2020} that performs a study focusing on software engineering in a service robotics perspective. However, those previous studies neither address SE in general for robotic systems nor SWEBOK \cite{SWEBOK} areas.

The \textbf{aim} for this study is to identify, classify and evaluate the software engineering systematic areas used in the robotic domain based on the SWEBOK guide. To achieve this goal, we systematically applied the mapping study \textbf{methodology} proposed by Kitchenham \al \cite{KITCHENHAM2013}.

The main \textbf{contributions} of this study are: (i) a systematic mapping of software engineering in robotic systems; (ii) a catalogue offering research studies about the utilization of software engineering in the robotic domain; and (iii) research opportunities and gaps in software engineering to robotic systems for future studies.
The \textbf{audience}  of this study is composed of both robotics researchers and roboticists interested in contributing to this research area and applying software engineering techniques to improve robotic systems.

Our study is organized as follows. Section \ref{sec:BACKGROUND} defines the concepts of robotic systems and SWEBOK guide. Section \ref{sec:RESEARCHMETHODOLOGY} describes in detail the adopted study design. Section \ref{sec:RESULTS} presents the results from our analysis. Section \ref{sec:DISCUSSION} discusses the main findings and the threats to validity. Section \ref{sec:RELATEDWORK} shows related studies. Section \ref{sec:CONCLUSIONS} synthesizes the final remarks and future work.

\section{BACKGROUND}
\label{sec:BACKGROUND}

From a software perspective, robotic systems have two layers: control and application layers. In the control layer, there are the drivers responsible for interacting directly with the hardware. The application layer is the software components responsible for defining robots' desired behaviour \cite{AHMAD2016}.

Typical characteristics of the robotic software are embedded, concurrent, real-time, distributed, data-intensive. Besides, robotics software needs to meet some quality requirements, such as quality aspects requirements, such as safety, reliability, and fault-tolerance \cite{SCHLEGEL2009}.
These characteristics and quality requirements are also present for software systems in other domains, such as avionics, automotive, factory automation, telecommunication and even large-scale information systems.

Software engineering is \textit{the systematic application of scientific and technological knowledge, methods, and experience to the design, implementation, testing, and documentation of software} \cite{ISO24765}. 
In 2004,  the first version of Guide to the Software Engineering Body of Knowledge (SWEBOK) was a mechanism for acquiring knowledge in software engineering professionals \cite{SWEBOK}. 

The SWEBOK is in version 3.0, and there are 15 Knowledge Areas (KA): Software Requirements, Software Design, Software Construction, Software Testing, Software Maintenance, Software Configuration Management, Software Engineering Management, Software Engineering Process, Software Engineering Models and Methods, Software Quality, Software Engineering Professional Practice, Software Engineering Economics, Computing Foundations, Mathematical Foundations, Engineering Foundations \cite{SWEBOK}.  
Although conventional and robotic systems are different, most studies about software engineering and robotics apply some SE techniques described in SWEBOK. For this reason, we decided to use it as the base for our classification framework.

\section{STUDY DESIGN}
\label{sec:RESEARCHMETHODOLOGY}

\begin{figure}[!h]
\centering
\includegraphics[width=.75\linewidth]{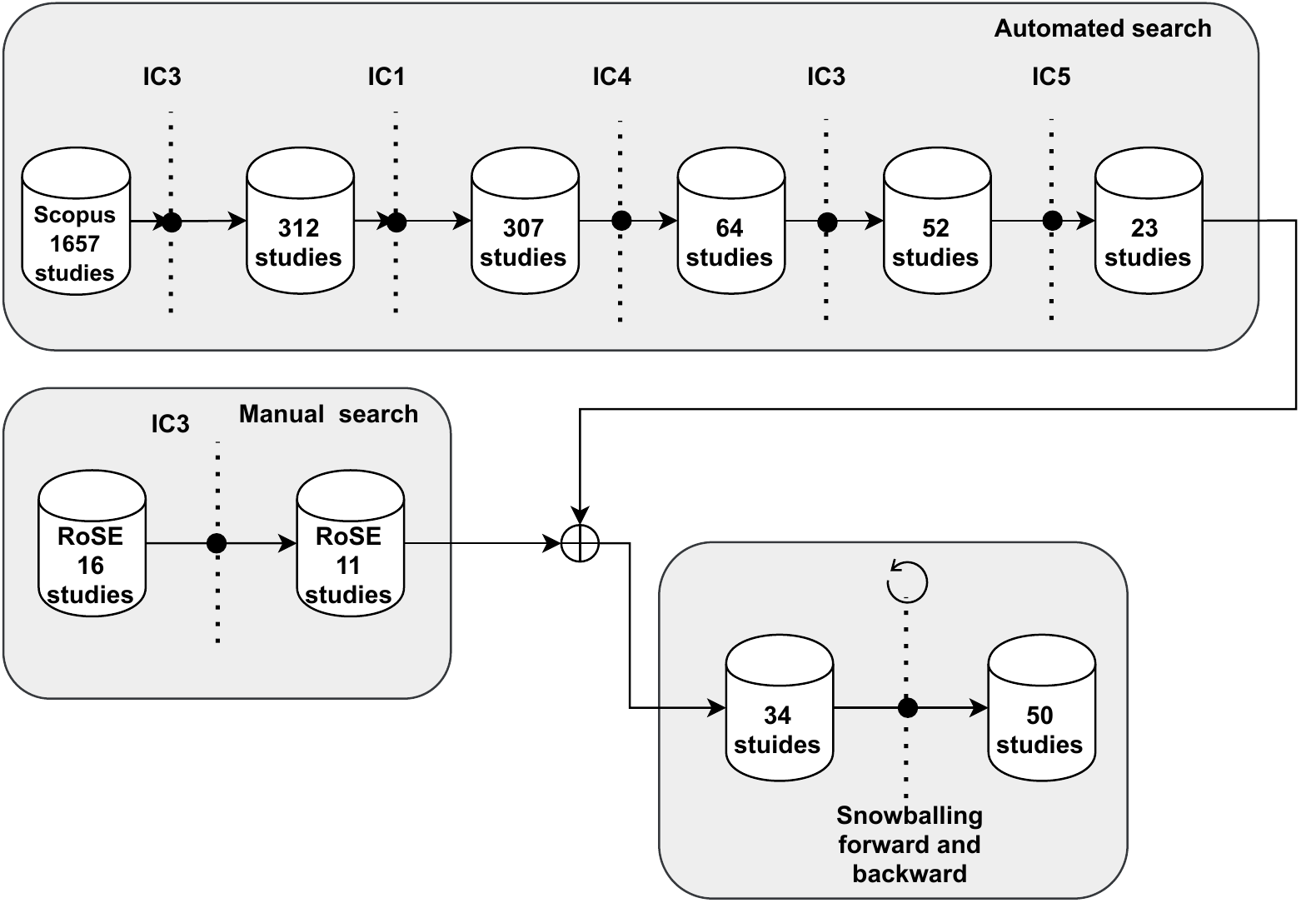}
\caption{Overview of the search strategy and study selection processes}
\label{fig:searchstrategy}
\end{figure}

The study followed the guidelines for systematic mapping studies \cite{KITCHENHAM2013}. Figure \ref{fig:searchstrategy} summarizes the steps of our study.

\subsection{Goal and Research Questions}

We refined our research goal into three research questions:

\textbf{RQ1: What are the demographics of publication of research studies about SE areas applied in the robotics domain?}
By answering this research question, we aim to understand how the research area is studied and relevant venues where researchers published their results on the topic.

\textbf{RQ2: What are the SWEBOK areas mentioned in the robotics domain?}
We studied the SWEBOK areas applied in robotics domains, including the 15 SWEBOK  areas defined in the guideline \cite{SWEBOK}.

\textbf{RQ3: What are the main findings of SERS?}
We aim to provide information to help researchers with evidence-based indications about the directions related to software engineering practices.

\subsection{Search strategy and study selection}

Following the systematic review process in software engineering guidelines, the first step is creating a  search query \cite{KITCHENHAM2007}. Our search query is: 

\begin{center} 
\texttt{((robot*)  AND  ("software engineering"))}
\end{center}

We defined the inclusion and exclusion criteria during the protocol definition to obtain a set of relevant studies to answer our RQs. We include a study if it satisfies all inclusion criteria and will discard it if the study met any exclusion criteria. 

\textbf{Inclusion criteria (IC)}

\begin{itemize}
\item \textbf{IC1:} The study must be a primary study;
 \item \textbf{IC2:} The paper must be an article or conference paper;
\item \textbf{IC3:} The study mentions software engineering topic in robotic domain;
\item \textbf{IC4:} The study was published in a venues that is ranked as A{*}, A and B for CORE Conference Ranking \cite{CORERANK};
\item \textbf{IC5:} The study must be a full paper.
\end{itemize}

\textbf{Exclusion criteria (EC)} 

\begin{itemize}
\item \textbf{EC1:} The study published as an abstract;
\item \textbf{EC2:} The study is not written in English;
\item \textbf{EC3:} The study is a short paper, keynote, tutorial, challenge, and showcase;
\item \textbf{EC4:} The  study  does  not  mention software engineering topic in robotic domain
\end{itemize}

In our study, we performed a hybrid search strategy proposed by Mourao \al \cite{MOURAO2020} that combine the search in a digital library and a parallel snowballing forward and backward. A forward snowballing refers to identifying new papers based on papers that cite citing the paper examined, and backward snowballing is the process to evaluate the references for the paper in the set \cite{WHOLIN2014}. 
Figure \ref{fig:searchstrategy} shows an overview of the search strategy and study selection that we performed in our study. A total of  1657  studies were turned from the automated search execution.  We removed conference announcements and applied the selection criteria  (IC  and  EC)  on titles and abstracts, leading to  23 candidate studies. Next, we combined the results of a manual search performed in the last editions of the International Workshop on Robotics Software Engineering (RoSE) after applying IC3, remaining in a single list with 34 studies. Finally, we performed forward and backward snowballing in the set of paper and with the results and applied exclusion and inclusion criteria on  title  and  abstract  leading to 16 studies that were added to the seed set of 34 studies. Remaining a initial final set of studies with 50 papers. The studies included in the review are listed in Table \ref{tab:primarystudies}.

\subsection{Data Extraction}

To extracted data from each primary study, we create a data extraction form. This form was created as a spreadsheet with specific fields to support our data extraction activity. 

First, we collected keywords and concepts through the full reading of the papers and applied the classification in a primary set of 10 papers. When the researchers found some divergence, they discussed, evaluated and refined the framework. The process ended when there were no primary studies to analyze left.

The classification framework is composed of three different categories, one for each research question.\\
\textbf{Publications Demographics (RQ1).} The parameters that we use collect data about publications trends are:  publication type and publication venue.\\
\textbf{Software Engineering areas in Robotics Domain (RQ2).} To answer research question 1 (RQ2), we used the software engineering  body of knowledge (SWEBOK). We are using \textit{SWEBOK Guideline} version 3 that is composed of 15 knowledge areas. \\
\textbf{Main findings of RSE (RQ3).} We highlight the results and conclusions of all studies evaluated. After that, we created clusters to conclude that they have similarities and described important aspects of robotics software engineering. We also eliminated the specific findings for one evaluated study because, in this case, the finding is not a global finding related to robotics software engineering.

\subsection{Data synthesis}
We summarize the extracted data with a combination of content analysis and narrative synthesis. We did not identify inconsistencies within the data. We present the data synthesis and our findings in Section \ref{sec:RESULTS} with a catalogue offering research studies about the utilization of software engineering in the robotic domain and in what SWEBOK Knowledge Area the study is concentrated (Table \ref{tab:primarystudies}).

\begin{table} 
\footnotesize
\caption{List of primary studies}
\centering
\label{tab:primarystudies}
\begin{tabular}{ C{0.75cm}  C{0.75cm}  L{10.5cm} C{2.5cm} }
\hline
\textbf{ID} & \textbf{Ref.} &\textbf{Title} &\textbf{SWEBOK Area} \\ 
\hline
\hline
\lightgrayrow
P2 &\cite{BAUML2006} &Agile Robot Development (aRD): A pragmatic approach to robotic software.& \cons\\ 
\silverrow
P7 &\cite{RiCKERT2017}&Robotics library: An object-oriented approach to robot applications.& \makecell{\cons} \\ 
\lightgrayrow
P9 &\cite{ZHONG2011}&Runtime models for automatic reorganization of multi-robot systems.&\cons \\ 
\silverrow
P14 &\cite{JAWAWI2007}&A component-oriented programming for embedded mobile robot software.&\makecell{\cons} \\
\lightgrayrow
P15 &\cite{GONZALEZ2013}&A new paradigm for open robotics research and education with the C++ OOML.&\cons \\
\silverrow
P24 &\cite{SHEPHERD2019}&Visualizing the Hidden ``Variables in Robot Programs''.&\makecell{\cons} \\
\lightgrayrow
P25 &\cite{GERASIMOU2019}&Towards Systematic Engineering of Collaborative Heterogeneous Robotic Systems.& \cons \\
\silverrow
P26 &\cite{RITSCHEL2019}&Novice-Friendly Multi-Armed Robotics Programming.&\cons \\
\lightgrayrow
P29 &\cite{MAOZ2018}&On the Software Engineering Challenges of Applying Reactive Synthesis to Robotics.&\makecell{\cons} \\
\silverrow
P32 &\cite{BELTRAME2018}&Engineering Safety in Swarm Robotics.&\makecell{\cons} \\
\lightgrayrow
P38 &\cite{KUDELSKI2013}& RoboNetSim: An integrated framework for multi-robot and network simulation.&\cons \\
\silverrow
P40 & \cite{SILVA2014}& Development of a flexible language for mission description for multi-robot missions.&\cons \\ 
\lightgrayrow
P45&\cite{WEINTROP2018}& Evaluating CoBlox: A comparative study of robotics programming environments for adult novices.& \cons \\
\silverrow
P46&\cite{NORDMANN2015}&Modeling of Movement Control Architectures based on Motion Primitives using Domain-Specific Languages&\cons\\
\lightgrayrow
P49&\cite{ALEXANDROVA2015}&RoboFlow: A flow-based visual programming language for mobile manipulation tasks&\cons\\
\silverrow
P18 &\cite{ZHANG2015}&Controlling Kuka industrial robots.&\makecell{\cons, \desi} \\
\lightgrayrow
P33 &\cite{ORE2018}&Towards Code-Aware Robotic Simulation: Vision Paper.&\makecell{\cons, \desi} \\
\silverrow
P35 & \cite{ALVAREZ2006} & An architectural framework for modeling teleoperated service robots.& \makecell{\cons, \desi}\\
\lightgrayrow
P37 & \cite{HAN2001}& Internet control architecture for internet-based personal robot.&\makecell{\cons, \desi}\\ 
\silverrow
P41 &\cite{PENG2016} &EmSBot A modular framework supporting the development of swarm robotics applications.&\makecell{\cons, \desi}\\ 
\lightgrayrow
P42 & \cite{SANCHEZ2011} & Introducing safety requirements traceability support in model-driven development of robotic applications.&\makecell{\cons, \desi} \\ 
\silverrow
P44 & \cite{SAUPPE014} & Design patterns for exploring and prototyping human-robot interactions.&\makecell{\cons, \desi} \\ 
\lightgrayrow
P20 &\cite{BRUGALI2018}&Software product line engineering for robotic perception systems.&\makecell{\cons, \mode} \\
\silverrow
P22 &\cite{AHN2012}&A framework-based approach for fault-tolerant service robots.&\makecell{\cons, \desi, \test} \\
\lightgrayrow
P1 & \cite{RAMASWAMY2014}& SafeRobots: A model-driven Framework for developing Robotic Systems.& \makecell{\cons, \desi, \mode} \\
\silverrow
P31 &\cite{BOZHINOSKI2018}&Designing Control Software for Robot Swarms: Software Engineering for the Development of Automatic Design Methods.&\makecell{\cons, \desi, \mode} \\
\lightgrayrow
P34 &\cite{ERNST2018}&Towards Rapid Composition with Confidence in Robotics Software.&\makecell{\cons, \quali, \test} \\
\silverrow
P21 &\cite{MOLINA2020}&Building the executive system of autonomous aerial robots using the Aerostack open-source framework.&\makecell{\cons, \quali, \desi} \\
\lightgrayrow
P3 & \cite{FLUCKIGER2014}& Service oriented robotic architecture for space robotics: Design, testing, and lessons learned.& \makecell{\desi}\\
\silverrow
P6 &\cite{TRIVINO2009}&Towards an architecture for semiautonomous robot telecontrol systems.& \makecell{\desi} \\
\lightgrayrow
P8 &\cite{KIM2006}&UML-based service robot software development: A case study. &\makecell{\desi} \\ 
\silverrow
P11 &\cite{MESSINA1999}&Component specifications for robotics integration.&\desi \\ 
\lightgrayrow
P17 &\cite{MINSEONG2009}&Service robot for the elderly: Software development with the COMET/UML method.&\desi \\
\silverrow
P27 &\cite{WITTE2018}&Checking Consistency of Robot Software Architectures in ROS.& \desi
\\
\lightgrayrow
P43 & \cite{PYO2015} & Service robot system with an informationally structured environment.&\desi \\
\silverrow
P47&\cite{KIM2005}&Re-engineering software architecture of home service robots: A case study&\desi\\
\lightgrayrow
P36 & \cite{MICHAEL2008}& Experimental Testbed for Large Multirobot Teams: Verification and Validation. & \desi \\
\silverrow
P4 &\cite{SILVA2012}&Designing a meta-model for a generic robotic agent system using Gaia methodology.&\makecell{\desi, \mode} \\ 
\lightgrayrow
P5 &\cite{AKIKI2020}&EUD-MARS: End-user development of model-driven adaptive robotics software systems.& \makecell{\desi, \mode} \\ 
\silverrow
P10 &\cite{PARACHOS2012}&Model-driven behavior specification for robotic teams.&\makecell{\desi, \mode} \\ 
\lightgrayrow
P13 &\cite{COEVOET2017}&Software toolkit for modeling, simulation, and control of soft robots.&\makecell{\desi, \mode} \\ 
\silverrow
P16 &\cite{ACERES2009}&Design of service robots: Experiences using software engineering.&\makecell{\desi, \mode} \\
\lightgrayrow
P19 &\cite{ORTIZ2015}&Model-driven analysis and design for software development of autonomous underwater vehicles.&\makecell{\desi, \mode} \\
\silverrow
P23 &\cite{MCKEE2001}&Object-oriented concepts for modular robotics systems.&\makecell{\desi, \mode}\\
\lightgrayrow
P28 &\cite{MILANO2018}&A Use Case in Model-Based Robot Development Using AADL and ROS.&\makecell{\desi, \mode} \\
\silverrow
P30 &\cite{BURGUEO2018}&Using Physical Quantities in Robot Software Models.&\makecell{\desi, \mode}\\
\lightgrayrow
P39 & \cite{CHO2012}& An interaction-driven approach to identifying functional behaviors of service robot systems.&\makecell{\desi, \test} \\ 
\silverrow
P12 &\cite{ABDELGAWAD2017}&Model-based testing of a real-time adaptive motion planning system.&\test \\ 
\lightgrayrow
P50&\cite{VONMAYRHAUSER1993}&Automated testing support for a robot tape library&\test \\
\silverrow
P48&\cite{CARLSON2003}&Reliability analysis of mobile robots&\makecell{\quali, \test}\\
\hline
\end{tabular}
\centering{
\footnotesize{SC= Software Construction, SD= Software Design, SMM= Software Models and Methods, SQ= Software Quality, \\ ST= Software Testing}}
\end{table}

\section{RESULTS} 
\label{sec:RESULTS}

\subsection{RQ1: What  are  the  demographics  of  publication  of research  studies  about  SE  areas  applied  in  the  robotics domain??}

We performed a hybrid search strategy that combines automated, manual and snowballing. The manual search was performed on the Robotics Software Engineering Workshop (RoSE) realized editions (2018 and 2019). For this reason, we have 11 studies from a workshop even though a workshop does not have the Core Rank of at least B. 

RoSE workshop studies are in our data set because it is a workshop focused on Robotics Software Engineering. Although the publication years are not a category that we use to answer the RQ1, the distribution in years is valuable to understand the importance of venues as RoSE. In 2018, which is the year of the first RoSE edition, we have the added of 8 studies, and in 2019 we have in our set of studies only studies were published only in the \textbf{RoSE}.

\textbf{Publications type} We classified the primary studies by type (conference or journal paper). The publication type distribution is 26 (11 from RoSE) studies published as a conference paper and 24 in journals. If we evaluated the set of studies without the papers from RoSE, we have 15 studies from conference papers and 24 from journals.

\textbf{Publications venues} The studies evaluated are distributed in 19 different venues (Table \ref{tab:tabvenues})  which research areas are Computer Science and Robotics. We observed a fragmentation in terms of publication venues, 10 of the venues have Robotics as the main research field, and nine are focused on Computer Science, and no venue addresses the intersection between robotics and software engineering. We observed that the significant number of studies published in one venue was 4, and it happened in journals whose main research field is robotics. Besides, three conferences ranked as A* in the set of venues, and we evaluated six studies published in these conferences.

\begin{table} [!htb]
\footnotesize
\caption{List of Venues}
\begin{tabular}{ L{7.5cm}  L{2cm}  C{2cm} C{3.5cm}} 
\hline
\textbf{Venue Title} & \textbf{Type} &\textbf{CORE Ranking }&{\centering}\textbf{Number of studies}
\\ \hline 
\silverrow
Information Sciences&Journal&A&3\\
Journal of Field Robotics&Journal&A&1\\
Science of Computer Programming&Journal&A&1\\
International Journal of Advanced Robotic Systems&Journal&B&4\\
IEEE Robotics and Automation Magazine&Journal&B&4\\
Autonomous Robots&Journal&B&3\\
Advanced Robotics&Journal&B&2\\
Robotica&Journal&B&2\\
International Journal of Semantic Computing&Journal&B&1\\
Robotics and Autonomous Systems&Journal&B&2\\
\lightgrayrow
International Conference on Software Engineering&Conference&A{*}&3  \\
\lightgrayrow
Conference on Human Factors in Computing Systems&Conference&A{*}&2\\
\lightgrayrow
International Conference on Autonomous Agents and Multi-agent Systems &Conference&A{*}&1 \\
\lightgrayrow
IEEE International Conference on Intelligent Robots and Systems &Conference&A&3 \\
\lightgrayrow
International Conference on Engineering of Complex Computer Systems&Conference&A&1\\
\lightgrayrow
International Symposium on Software Reliability Engineering&Conference&A&1\\ 
\lightgrayrow
IEEE International Conference on Robotics and Automation&Conference&B&3\\
\lightgrayrow
Technology of Object-Oriented Languages and Systems&Conference&B&1\\
\hline
\end{tabular}
\label{tab:tabvenues}
\end{table}

\begin{tcolorbox}[float, colframe=gray!25, coltitle=black, arc=0mm, title=\textbf{Observation 1 - Demographics of publication}]
There is a  distribution in venues with two different research areas, \textbf{Computer Science and Robotics}. We observed that RoSE play a critical role to the development for a interdisciplinary research field as Robotics Software Engineering, besides the studies are published in journals more than in conferences. \end{tcolorbox}

\subsection{RQ2: What  are  the  SWEBOK  areas  mentioned  in  the robotics domain?}

We aim with RQ2 to provide an overview for researchers and practitioners about the established software engineering areas studied in robotic domains. As shown in Table \ref{tab:tabswebokpapers}, 5 of 15 SWEBOK knowledge areas appear on the studies evaluated for us.

\begin{table} [!htb]
\footnotesize
\centering
\caption{List of included studies for each SWEBOK knowledge areas}
\begin{tabular}{ L{3.5cm}  L{7cm}  C{3.5cm} } 
\hline
\textbf{Knowledge Area} & \textbf{References} &{\centering}\textbf{Number of studies}\\ \hline \\
\textbf{Software Design}&P1, P3, P4, P5, P6, P8, P10, P11, P12, P13, P16, P17, P18, P19, P21, P22, P23, P27, P28, P30, P31, P33, P35, P36,  P37, P39, P41, P42, P43,  P44, P47 &31\\ \\
\textbf{Software Construction}&P1, P2, P6, P7, P9, P14, P15, P18, P20, P21, P22, P24, P25, P26, P29, P31, P32, P33, P34, P35, P38, P40, P41, P42, P44, P45, P46, P49 &27\\ \\
\textbf{Software Models and Methods} &P1, P4, P5, P10, P13, P16, P19, P20, P23, P28, P30, P31  &12\\ \\
\textbf{Software Testing}&P12, P22,  P34, P39, P48, P50 &6 \\ \\
\textbf{Software Quality}&P21, P34, P48&3\\
\hline
\end{tabular}
\label{tab:tabswebokpapers}
\end{table}

The SWEBOK areas was the framework used to classify the studies. Thus, the SWEBOK is more cited in our set of studies was \textit{Software Design} with 60\% of the studies, and the techniques applied in the software design area are UML, model-driven development, component-based software engineering, architecture and description language.

The second SWEBOK area in our rank was \textit{Software Construction} (54\%), with most of the studies proposing a framework for solving software problems in the robotics domain.

\textit{Software Models and Methods} with 24\% of the studies is the third SWEBOK area. The SE techniques applied in this area are the meta-model, analysis of models, framework model. 

\textit{Software Testing} (12\%)  and \textit{Software Quality} (6\%) are the two of five SWEBOK areas cited with fewer studies evaluated. The SE techniques investigated are test generation method, code quality standard, requirement analysis and a fault-tolerance architecture.

\begin{figure}[!htb]
\centering
\includegraphics[width=.65\linewidth]{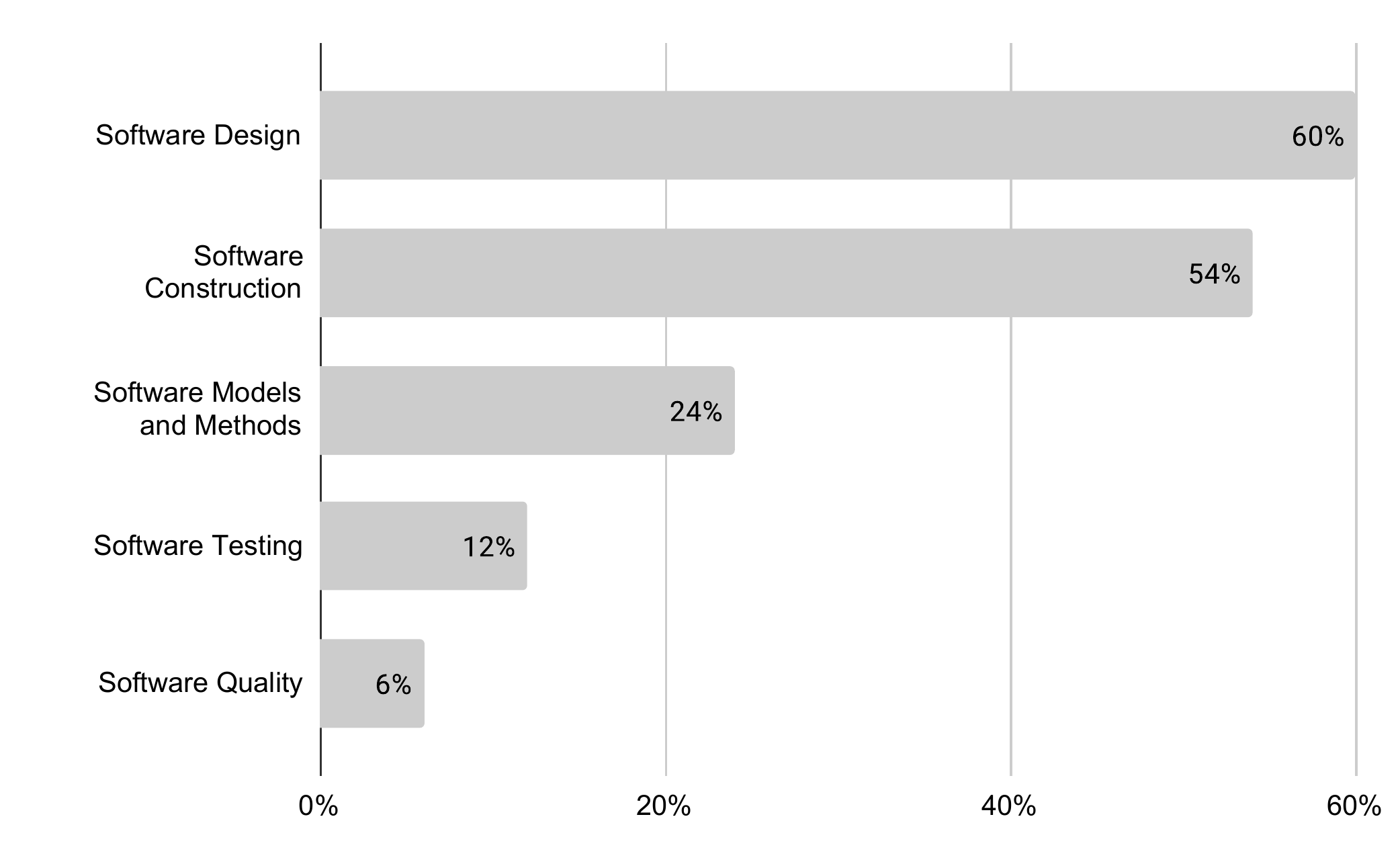} 
\caption{Number of studies and SE technique applied in Robotics Domain.}
\label{fig:resultsrq2}
\end{figure}

\begin{tcolorbox}[float, colframe=gray!25, coltitle=black, arc=0mm, title=\textbf{Observation 2 - SWEBOK areas in Robotics domain}]
The majority of the studies are concentrated in 3 areas that performed activities inherent to robotics software development (\textit{software design}, \textit{construction} and \textit{models and methods}) . Only 18\% of the studies performed activities related to testing and quality, and the evaluated studies do not cover 10 SWEBOK areas, as shown in Figure \ref{fig:resultsrq2}. \end{tcolorbox}
 
\subsection{RQ3: What  are  the  main  findings  of  SERS?}

With the answers for RQ3, we aim to provide information to help researchers with evidence-based indications about the directions related to software engineering practices that are used.

The first finding is related to \textbf{the evidence in our set of studies that software engineering practices improve robotic systems quality.} According to Aceres \al \cite{ACERES2009}, software engineering contributes to the quality improvement of robotic systems and reduces the effort related to the development. Besides, in the study performed by Kim \al \cite{KIM2006}, it is highlighted that it is impossible to resolve integration and developing issues in robotic systems without systematic and comprehensive software development method. Finally, the standard development is described as a key to the usability and sustainability of the robotic assets \cite{FLUCKIGER2014}.

Another important finding is \textbf{how software engineering practices can help disseminate development for robotic systems}. There is evidence that applying approaches based on model-driven development can empower not professional programmers to develop robotics software without requiring advanced technical skills\cite{AKIKI2020}. In the study performed by Gerasimou \al \cite{GERASIMOU2019}, the Domain Specific Language concepts allow both software engineers and robotic experts to write robots software with more facility. Gonzales performed a study that allows engineers to create robotic platforms or other physical designs to exchange and collaborate with other researchers. \cite{GONZALEZ2013}.  

\begin{tcolorbox}[float, colframe=gray!25, coltitle=black, arc=0mm, title=\textbf{Observation 3 - Main Findings about SER}]
There is a consensus by the research community that software engineering techniques and approaches can improve the quality of robot software. 
\end{tcolorbox}

\section{DISCUSSIONS}
\label{sec:DISCUSSION}

In this study, we evaluated Robotics Software Engineering to elicit how the SE areas are applied in the robotic domains and the gaps/opportunities for future work in the intersection of these two research fields. 

We highlight two aspects of the study's demographics. First, we observed a \textbf{distribution related to the publications venues} (Table \ref{tab:tabvenues}), the evaluated studies are published in 19 different venues. The studies are published in venues where the main research field is Robotics (9) or Computer Science(9), further the workshop RoSE. Second, most journals focus on robotics, and most conferences focus on computer science. The venues ranked in the highest level are three conferences: International Conference on Software Engineering Conference, Conference on Human Factors in Computing Systems Conference, International Conference on Autonomous Agents and Multi-agent Systems Conference. 

These facts lead us to conclude that there is not a concentration in a unique venue or area. Moreover, through observation, a dissonance between the methodology applied in the evaluated studies published in robotic venues concerned software engineering techniques and approaches. For example, in the studies that use model-driven development,  80\% do not have empirical study as is trivial in a software engineering study. 

In the context of SE Areas, we observed that the evaluated studies cited 5/15 SWEBOK Knowledge Areas: \textit{Software Design, Software Construction, Software Models and Methods, Software Testing and Software Quality}. The top 3 areas mentioned are related to the Design and Implementation phase of the traditional approach of system development: \textit{Software Design, Software Construction and Software Models \& Methods}. However, 10/15 SE areas are not mentioned, and 5 of these studies are deeply researched by the software engineering community: \textit{Software Requirements, Software Maintenance, Software Configuration Management,  Software Engineering  Management,  Software Engineering  Process}. We believe that there are gaps and research opportunities in these 5 SE areas for robotic domains.  

The studies that cited topics related to the \textit{Software Models and Methods} area also applied some topics related to the Software Design area (11/29). Besides, only five studies (P5, P10, P19, P20, P28) in \textit{Software Models and Methods}  declare they are working with Model-driven development (MDD). However, 80\%  of the studies related to MDD have evaluation issues. For instance, in the study performed by \cite{ACERES2009}, it is unclear how the authors evaluated the part of the proposal related to MDD. This observation suggests a \textbf{lack of empirical evaluation in the projects that apply MDD}.

\textit{Software construction} is the second area more cited in the evaluated studies. According to the SWEBOK \cite{SWEBOK}, software construction is a SE area related to coding, verification, unit testing, integration testing, and debugging. However, in the studies evaluated, the technique more applied is software construction tools with a focus on coding as the studies performed by Rickert \al \cite{RiCKERT2017} that describe a library approach named Robotics Library (RL), and by Alvarez \al \cite{ALVAREZ2006} which purpose is presenting a reference architectural framework for service robot control applications.    

\textit{Software testing} was cited only in 12\% of the studies. This result corroborated with other studies, for example, in the survey about testing robotic system performed Afzal \cite{AFZAL2020} \al. In this study, the author mentioned that robots interact with a physical component, which is a challenge for testing robotic systems.

Besides, only 6\% of the studies evaluated are concerned \textit{Software Quality}. We highlight that for our classification, it was important not only to cite quality requirements but also to develop topics related to \textit{Software Quality} topics such as software quality fundamentals, software quality management processes, practical considerations, and software quality tools \cite{FAIRLEY2014}.

Threats to validity usually happen in a mapping study, and it was not different in our study. We highlight some of those threats and the mechanism that we applied to address them. 

First, the main limitation is that our search string is restrictive concerning software engineering terms.  This choice could lead us to primary studies that are not representative of state-of-the-art robotics software engineering. We mitigate this threat by applying a hybrid search strategy proposed by Mourao \al, and Wholin \al \cite{MOURAO2020, WHOLIN2014}, with automated and manual search in the first stage and backward-forward snowballing. 

Second, we considered only peer-reviewed studies published in venues with CORE Conference Ranking at least B \cite{CORERANK}; this potential bias did not impact our study since the studies evaluated were reviewed in a rigorous process, which is a requirement for high-quality.

Another limitation of this work is the established categories. We could identify other categories if we decided, for example, to classify the approaches and techniques.  However, we opted to follow the SWEBOK guideline because it is a known guideline used in SE. Besides, we could answer other RQs and provide a broad overview of SE for robotic systems. 

\section{RELATED WORK}
\label{sec:RELATEDWORK}

Ahmad \al \cite{AHMAD2016} performed a systematic mapping study on software architectures for robotic systems on a set of  56 primary studies. The authors identified eight themes that support architectural solutions, the progress from object-oriented to component-based, and cloud robotics as an emerging solution. Our study differs from \cite{AHMAD2016} in the following terms: (i) we apply a more general search process by considering studies related to all SE areas, so we use a search string with the expression \texttt{"software engineering"} instead of using specific terms, for example, architecture, framework or development; (ii) we use a known guideline, \textit{SWEBOK guide} \cite{SWEBOK}, as the framework to classify the studies. 

In \cite{BOLZHINOSKI2019}, Bozhinoski \al performed a systematic mapping study on safety for mobile robotic systems with a software engineering perspective. They analyzed 58 primary studies. Their study focuses on existing solutions that use software engineering methods or techniques to manage mobile robotic systems' safety. Our study does not analyse solutions in a specific type of robotic system. For this reason, we use the term \texttt{robotics*} in our search strategy/ Alos, our study aims to analyze and identify all the studies that apply \textit{Software Quality} techniques, as software quality fundamentals, software quality management processes, practical consideration, software quality tools \cite{FAIRLEY2014}.

Malavolta \al \cite{MALAVOLTA2020} performed a study to elicit evidence-based architectural guidelines for open-source ROS-based software for robots. They constructed a data-set of 335 GitHub repositories containing real ROS-based systems and an online survey. Their study plays an important role in the mapping of Software Engineering for Robotics, given the increase of the ROS utilization by roboticists and architecture robotic researchers. Our study differs from their study because (i) we apply an empirical method, i.e., systematic mapping; (ii) we do not focus on a specific middleware (ROS). 

Garcia \cite{GARCIA2020} have \al reported a survey study of robotics software engineering from a service robotics domains perspective. The survey evaluated the robotics system's state through 18 semi-structured interviews with industrial practitioners and a survey with 156 respondents. The authors provide a comprehensive picture of software engineering practices in the robotics domain, specific characteristics of robotics software engineering and the usual challenges with adopted solutions. The differences between the work by Garcia \al and our study is: (i) we apply a different methodology in our study, i.e., we apply a systematic mapping using the guideline proposed by \cite{KITCHENHAM2007}; (ii) we do not focus on a type of robots; (iii) our study has data from academic publications. We considerate our study and the studies performed by Malavolta \cite{MALAVOLTA2020} \al, and by Garcia \cite{GARCIA2020} \al as complementary studies about robotics software engineering.

\section{CONCLUSIONS}
\label{sec:CONCLUSIONS}

We evaluated 50 primary studies and produced a systematic mapping on SE areas in the robotics domain based on the SWEBOK guide, and we identify gaps to lead research in the robotics software development. Nineteen different venues(journals and conferences) concentrated on the studies evaluated. The majority of the journals focus on Robotics, and the venues with a higher level in the Core Ranking are Computer Science conferences. We evaluated 24 journal studies,  15 conference studies and 11 studies from the RoSE workshop.

We observe that the academy's effort to apply SE techniques with 5/15 SWEBOK areas cited in the studies in the last years. However, most of the studies (Software design 60\%, Software Construction 54\% and Software Models and Methods 24\%) are related to the development phase. Other areas, for example, software testing (12\%) and quality(4\%), are not broadly applied in robotic domains or published. 

Besides, there are ten areas in the SWEBOK that are not mentioned in the evaluated studies. We speculate that there are 5/10 SE area that is not cited but can significantly impact robotic systems' software development. We believe that there are SE areas not properly established yet in robotic domains, a research opportunity that needs to be addressed by research robotic and software engineering.

Robotic software development focuses on integrating and reusable components \cite{GARCIA2020}, so model-driven and component-based using models, methods, and frameworks to construct software solutions are broadly applied. However, through our study, we observed a gap in empirical studies that evaluated the utilization of model-driven robotic domains.

We hope with our study to contribute to the movement of synergy between robotics and software engineering communities. Based on our discussion, we believe that there is a gap in software testing and quality for robotics systems,  and the community must focus on these SE areas to improve robotics software components. As future work, we intend to perform our research query in three other software engineering digital libraries: ACM Digital Library, IEEE Xplore, and Web of Science. Besides that, we aim to perform a cross-analysis among our study and surveys presented in the related works \cite{GARCIA2020, MALAVOLTA2020}.

\bibliographystyle{unsrt}
\bibliography{main}

\begin{thebibliography}{10}

\bibitem{AHMAD2016}
Aakash Ahmad and Muhammad~Ali Babar.
\newblock {Software architectures for robotic systems: A systematic mapping
  study}.
\newblock {\em Journal of Systems and Software}, 122:16--39, 2016.

\bibitem{ALBUQUERQUE017}
Danyllo Albuquerque, J.~Castro, Sarah Ribeiro, and T.~Heineck.
\newblock {Requirements Engineering for Robotic System: A Systematic Mapping
  Study}.
\newblock {\em WER}, 2017.

\bibitem{MALAVOLTA2020}
Ivano Malavolta, Grace~A Lewis, Bradley Schmerl, Patricia Lago, and David
  Garlan.
\newblock {How do you Architect your Robots? State of the Practice and
  Guidelines for ROS-based System}.
\newblock In {\em Proceedings of the 42nd International Conference on Software
  Engineering: Software Engineering in Practice}, 2020.

\bibitem{GARCIA2020}
Sergio Garc\'{\i}a, Daniel Str\"{u}ber, Davide Brugali, Thorsten Berger, and
  Patrizio Pelliccione.
\newblock {\em Robotics Software Engineering: A Perspective from the Service
  Robotics Domain}, page 593–604.
\newblock Association for Computing Machinery, New York, NY, USA, 2020.

\bibitem{SWEBOK}
Pierre Bourque, Richard~E. Fairley, and IEEE~Computer Society.
\newblock {\em Guide to the Software Engineering Body of Knowledge (SWEBOK(R)):
  Version 3.0}.
\newblock IEEE Computer Society Press, Washington, DC, USA, 3rd edition, 2014.

\bibitem{KITCHENHAM2013}
Barbara Kitchenham and Pearl Brereton.
\newblock A systematic review of systematic review process research in software
  engineering.
\newblock {\em Inf. Softw. Technol.}, 55(12):2049--2075, December 2013.

\bibitem{SCHLEGEL2009}
C.~{Schlegel}, T.~{Hassler}, A.~{Lotz}, and A.~{Steck}.
\newblock Robotic software systems: From code-driven to model-driven designs.
\newblock In {\em 2009 International Conference on Advanced Robotics}, pages
  1--8, 2009.

\bibitem{ISO24765}
{Systems and software engineering — Vocabulary}, {2017}.

\bibitem{KITCHENHAM2007}
Kitchenham BA and Stuart Charters.
\newblock Guidelines for performing systematic literature reviews in software
  engineering.
\newblock 2, 01 2007.

\bibitem{CORERANK}
Computing Research~\& Education.
\newblock Core rankings portal.

\bibitem{MOURAO2020}
Erica Mourão, João~Felipe Pimentel, Leonardo Murta, Marcos Kalinowski, Emilia
  Mendes, and Claes Wohlin.
\newblock On the performance of hybrid search strategies for systematic
  literature reviews in software engineering.
\newblock {\em Information and Software Technology}, 123:106294, 2020.

\bibitem{WHOLIN2014}
Claes Wohlin.
\newblock Guidelines for snowballing in systematic literature studies and a
  replication in software engineering.
\newblock In {\em Proceedings of the 18th International Conference on
  Evaluation and Assessment in Software Engineering}, EASE '14, New York, NY,
  USA, 2014. Association for Computing Machinery.

\bibitem{BAUML2006}
Berthold B{\"{a}}uml and Gerd Hirzinger.
\newblock {Agile Robot Development (aRD): A pragmatic approach to robotic
  software}.
\newblock {\em IEEE International Conference on Intelligent Robots and
  Systems}, pages 3741--3748, 2006.

\bibitem{RiCKERT2017}
Markus Rickert and Andre Gaschler.
\newblock {Robotics Library : An Object-Oriented Approach to Robot
  Applications}.
\newblock 2017.

\bibitem{ZHONG2011}
Christopher Zhong, Nichols Hall, Scott~A Deloach, and Nichols Hall.
\newblock {Runtime Models for Automatic Reorganization of Multi-Robot Systems}.
\newblock pages 20--29, 2011.

\bibitem{JAWAWI2007}
Dayang~N.A. Jawawi, Rosbi Mamat, and Safaai Deris.
\newblock {A component-oriented programming for embedded mobile robot
  software}.
\newblock {\em International Journal of Advanced Robotic Systems},
  4(3):371--380, 2007.

\bibitem{GONZALEZ2013}
Alberto Valero-g{\'{o}}mez~Juan Gonz{\'{a}}lez-g{\'{o}}mez.
\newblock {A new paradigm for open robotics research and education with the C
  ++ OOML}.
\newblock pages 233--249, 2013.

\bibitem{SHEPHERD2019}
David~C. Shepherd, Nicholas~A. Kraft, and Patrick Francis.
\newblock {Visualizing the 'hidden' variables in robot programs}.
\newblock {\em Proceedings - 2019 IEEE/ACM 2nd International Workshop on
  Robotics Software Engineering, RoSE 2019}, pages 13--16, 2019.

\bibitem{GERASIMOU2019}
Simos Gerasimou, Nicholas Matragkas, and Radu Calinescu.
\newblock {Towards systematic engineering of collaborative heterogeneous
  robotic systems}.
\newblock {\em Proceedings - 2019 IEEE/ACM 2nd International Workshop on
  Robotics Software Engineering, RoSE 2019}, pages 25--28, 2019.

\bibitem{RITSCHEL2019}
Nico Ritschel, Reid Holmes, Ronald Garcia, and David Shepherd.
\newblock {Novice-friendly multi-armed robotics programming}.
\newblock {\em Proceedings - 2019 IEEE/ACM 2nd International Workshop on
  Robotics Software Engineering, RoSE 2019}, pages 29--32, 2019.

\bibitem{MAOZ2018}
Shahar Maoz and Jan~Oliver Ringert.
\newblock {On the software engineering challenges of applying reactive
  synthesis to robotics}.
\newblock {\em Proceedings - International Conference on Software Engineering},
  pages 17--22, 2018.

\bibitem{BELTRAME2018}
Giovanni Beltrame, Ettore Merlo, Jacopo Panerati, and Carlo Pinciroli.
\newblock {Engineering safety in swarm robotics}.
\newblock {\em Proceedings - International Conference on Software Engineering},
  pages 36--39, 2018.

\bibitem{KUDELSKI2013}
Michal Kudelski, Luca~M. Gambardella, and Gianni~A. {Di Caro}.
\newblock {RoboNetSim: An integrated framework for multi-robot and network
  simulation}.
\newblock {\em Robotics and Autonomous Systems}, 61(5):483--496, 2013.

\bibitem{SILVA2014}
Daniel~Castro Silva, Pedro~Henriques Abreu, Lu{\'{i}}s~Paulo Reis, and
  Eug{\'{e}}nio Oliveira.
\newblock {Development of a flexible language for mission description for
  multi-robot missions}.
\newblock {\em Information Sciences}, 288(1):27--44, 2014.

\bibitem{WEINTROP2018}
David Weintrop, Afsoon Afzal, Jean Salac, Patrick Francis, Boyang Li, David
  Shepherd, and Diana Franklin.
\newblock {Evaluating CoBlox}.
\newblock (February 2019):1--1, 2018.

\bibitem{NORDMANN2015}
Arne Nordmann, Sebastian Wrede, and Jochen Steil.
\newblock {Modeling of movement control architectures based on motion
  primitives using domain-specific languages}.
\newblock {\em Proceedings - IEEE International Conference on Robotics and
  Automation}, 2015-June(June):5032--5039, 2015.

\bibitem{ALEXANDROVA2015}
Sonya Alexandrova, Zachary Tatlock, and Maya Cakmak.
\newblock {RoboFlow: A flow-based visual programming language for mobile
  manipulation tasks}.
\newblock {\em Proceedings - IEEE International Conference on Robotics and
  Automation}, 2015-June(June):5537--5544, 2015.

\bibitem{ZHANG2015}
Houxiang Zhang and Massimiliano Fago.
\newblock {Controlling Kuka Industrial Robots: Flexible Communication Interface
  JOpenShowVar}.
\newblock (December), 2015.

\bibitem{ORE2018}
John-paul Ore, Carrick Detweiler, and Sebastian Elbaum.
\newblock {Towards Code-Aware Robotic Simulation Vision Paper}.
\newblock pages 40--43, 2018.

\bibitem{ALVAREZ2006}
{An architectural framework for modeling teleoperated service robots}.
\newblock {\em Robotica}, 24(4):411--418, 2006.

\bibitem{HAN2001}
Kuk~Hyun Han, Sinn Kim, Yong~Jae Kim, and Jong~Hwan Kim.
\newblock {Internet control architecture for internet-based personal robot}.
\newblock {\em Autonomous Robots}, 10(2):135--147, 2001.

\bibitem{PENG2016}
Long Peng, Fei Guan, Luc Perneel, and Martin Timmerman.
\newblock {EmSBot A modular framework supporting the development of swarm
  robotics applications}.
\newblock {\em International Journal of Advanced Robotic Systems}, 13(6):1--15,
  2016.

\bibitem{SANCHEZ2011}
Pedro S{\'{a}}nchez, Diego Alonso, Francisca Rosique, B{\'{a}}rbara
  {\'{A}}lvarez, and Juan~A. Pastor.
\newblock {Introducing safety requirements traceability support in model-driven
  development of robotic applications}.
\newblock {\em IEEE Transactions on Computers}, 60(8):1059--1071, 2011.

\bibitem{SAUPPE014}
Allison Saupp{\'{e}} and Bilge Mutlu.
\newblock {Design patterns for exploring and prototyping human-robot
  interactions}.
\newblock {\em Conference on Human Factors in Computing Systems - Proceedings},
  pages 1439--1448, 2014.

\bibitem{BRUGALI2018}
Davide Brugali and Nico Hochgeschwender.
\newblock Software product line engineering for robotic perception systems.
\newblock {\em International Journal of Semantic Computing}, 12:89--107, 03
  2018.

\bibitem{AHN2012}
Heejune Ahn, Woong~Kee Loh, and Woon~Young Yeo.
\newblock {A framework-based approach for fault-tolerant service robots}.
\newblock {\em International Journal of Advanced Robotic Systems}, 9:1--10,
  2012.

\bibitem{RAMASWAMY2014}
Arunkumar Ramaswamy, Bruno Monsuez, and Adriana Tapus.
\newblock {SafeRobots: A model-driven Framework for developing Robotic
  Systems}.
\newblock {\em IEEE International Conference on Intelligent Robots and
  Systems}, (Iros):1517--1524, 2014.

\bibitem{BOZHINOSKI2018}
Darko Bozhinoski, Universit{\'{e}} Libre, Mauro Birattari, and Universit{\'{e}}
  Libre.
\newblock {Designing Control Software for Robot Swarms: Software Engineering
  for the Development of Automatic Design Methods}.
\newblock pages 33--35, 2018.

\bibitem{ERNST2018}
Neil~A Ernst, Rick Kazman, and Philip Bianco.
\newblock {Towards Rapid Composition with Confidence in Robotics Software}.
\newblock pages 44--47, 2018.

\bibitem{MOLINA2020}
Martin Molina, Abraham Carrera, Alberto Camporredondo, Hriday Bavle, and
  Alejandro Rodriguez-ramos.
\newblock {Building the executive system of autonomous aerial robots using the
  Aerostack open-source framework}.
\newblock (June):1--20, 2020.

\bibitem{FLUCKIGER2014}
Lorenzo Fl{\"{u}}ckiger and Hans Utz.
\newblock {Service oriented robotic architecture for space robotics: Design,
  testing, and lessons learned}.
\newblock {\em Journal of Field Robotics}, 31(1):176--191, 2014.

\bibitem{TRIVINO2009}
Gracian Trivino, Luis Mengual, and Albert van~der Heide.
\newblock {Towards an architecture for semiautonomous robot telecontrol
  systems}.
\newblock {\em Information Sciences}, 179(23):3973--3984, 2009.

\bibitem{KIM2006}
Minseong Kim, Suntae Kim, Sooyong Park, Mun~Taek Choi, Munsang Kim, and Hassan
  Gomaa.
\newblock {UML-based service robot software development: A case study}.
\newblock {\em Proceedings - International Conference on Software Engineering},
  2006(Icse):534--543, 2006.

\bibitem{MESSINA1999}
Elena Messina, John Horst, Thomas Kramer, Hui~Min Huang, and John Michaloski.
\newblock {Component specifications for robotics integration}.
\newblock {\em Autonomous Robots}, 6(3):247--264, 1999.

\bibitem{MINSEONG2009}
Kim Minseong, Kim Suntae, Park Sooyong, Choi Mun-Taek, Kim Munsang, and Hassan
  Gomaa.
\newblock {Service robot for the elderly: Software development with the
  COMET/UML method}.
\newblock {\em IEEE Robotics and Automation Magazine}, 16(1):34--45, 2009.

\bibitem{WITTE2018}
Thomas Witte and Matthias Tichy.
\newblock {Checking consistency of robot software architectures in ROS}.
\newblock {\em Proceedings - International Conference on Software Engineering},
  pages 1--8, 2018.

\bibitem{PYO2015}
Yoonseok Pyo, Kouhei Nakashima, Shunya Kuwahata, Ryo Kurazume, Tokuo Tsuji,
  Ken'ichi Morooka, and Tsutomu Hasegawa.
\newblock {Service robot system with an informationally structured
  environment}.
\newblock {\em Robotics and Autonomous Systems}, 74:148--165, 2015.

\bibitem{KIM2005}
Moonzoo Kim, Jaejoon Lee, Kyo~Chul Kang, Youngjin Hong, and Seokwon Bang.
\newblock {Re-engineering software architecture of home service robots: A case
  study}.
\newblock {\em Proceedings - 27th International Conference on Software
  Engineering, ICSE05}, pages 505--513, 2005.

\bibitem{MICHAEL2008}
Nathan Michael, Jonathan Fink, and Vijay Kumar.
\newblock {Experimental Testbed for Large Multirobot Teams: Verification and
  Validation}.
\newblock {\em IEEE Robotics and Automation Magazine}, 15(1):53--61, 2008.

\bibitem{SILVA2012}
Daniel~Castro Silva, Rodrigo~A.M. Braga, Lu{\'{i}}s~Paulo Reis, and
  Eug{\'{e}}nio Oliveira.
\newblock {Designing a meta-model for a generic robotic agent system using Gaia
  methodology}.
\newblock {\em Information Sciences}, 195:190--210, 2012.

\bibitem{AKIKI2020}
Pierre~A. Akiki, Paul~A. Akiki, Arosha~K. Bandara, and Yijun Yu.
\newblock {EUD-MARS: End-user development of model-driven adaptive robotics
  software systems}.
\newblock {\em Science of Computer Programming}, 200:102534, 2020.

\bibitem{PARACHOS2012}
Alexandras Paraschos, Nikolaos~I. Spanoudakis, and Michail~G. Lagoudakis.
\newblock {Model-driven behavior specification for robotic teams}.
\newblock {\em 11th International Conference on Autonomous Agents and
  Multiagent Systems 2012, AAMAS 2012: Innovative Applications Track}, 2(March
  2014):696--703, 2012.

\bibitem{COEVOET2017}
E.~Coevoet, T.~Morales-Bieze, F.~Largilliere, Z.~Zhang, M.~Thieffry,
  M.~Sanz-Lopez, B.~Carrez, D.~Marchal, O.~Goury, J.~Dequidt, and C.~Duriez.
\newblock {Software toolkit for modeling, simulation, and control of soft
  robots}.
\newblock {\em Advanced Robotics}, 31(22):1208--1224, 2017.

\bibitem{ACERES2009}
Diego Alonso~C Aceres, Francisco~J Ortiz, Juan~Pastor Franco, and Pedro~S
  Anchez.
\newblock {Design of Service Robots}.
\newblock (March), 2009.

\bibitem{ORTIZ2015}
{Model-driven analysis and design for software development of autonomous
  underwater vehicles}.
\newblock {\em Robotica}, 33(8):1731--1750, 2015.

\bibitem{MCKEE2001}
G.~T. McKee, J.~A. Fryer, and P.~S. Schenker.
\newblock {Object-oriented concepts for modular robotics systems}.
\newblock {\em "Technology of Object-Oriented Languages and Systems},
  (TOOL):229--239, 2001.

\bibitem{MILANO2018}
Politecnico Milano, Andrea Semprebon, Politecnico Milano, Matteo Matteucci, and
  Politecnico Milano.
\newblock {A use case in model-based robot development using AADL and ROS
  Gianluca Bardaro}.
\newblock pages 9--16, 2018.

\bibitem{BURGUEO2018}
Loli BurguEo, Tanja Mayerhofer, Manuel Wimmer, and Antonio Vallecillo.
\newblock {Using physical quantities in robot software models}.
\newblock {\em Proceedings - International Conference on Software Engineering},
  pages 23--28, 2018.

\bibitem{CHO2012}
Youngdo Cho, Hwangwook Kim, Dae~Kyoo Kim, and Sooyong Park.
\newblock {An interaction-driven approach to identifying functional behaviors
  of service robot systems}.
\newblock {\em Proceedings - 2012 IEEE 17th International Conference on
  Engineering of Complex Computer Systems, ICECCS 2012}, pages 109--118, 2012.

\bibitem{ABDELGAWAD2017}
Mahmoud Abdelgawad, Sterling McLeod, Anneliese Andrews, and Jing Xiao.
\newblock {Model-based testing of a real-time adaptive motion planning system}.
\newblock {\em Advanced Robotics}, 31(22):1159--1176, 2017.

\bibitem{VONMAYRHAUSER1993}
Anneliese {Von Mayrhauser} and Stewart Crawford-Hines.
\newblock {Automated testing support for a robot tape library}.
\newblock {\em Proceedings - International Symposium on Software Reliability
  Engineering, ISSRE}, pages 6--14, 1993.

\bibitem{CARLSON2003}
Jennifer Carlson and Robin~R. Murphy.
\newblock {Reliability analysis of mobile robots}.
\newblock {\em Proceedings - IEEE International Conference on Robotics and
  Automation}, 1:274--281, 2003.

\bibitem{AFZAL2020}
Afsoon Afzal, Claire~Le Goues, Michael Hilton, and Christopher~Steven
  Timperley.
\newblock {A Study on Challenges of Testing Robotic Systems}.
\newblock {\em Proceedings - 2020 IEEE 13th International Conference on
  Software Testing, Verification and Validation, ICST 2020}, pages 96--107,
  2020.

\bibitem{FAIRLEY2014}
R.~E.~D. {Fairley}, P.~{Bourque}, and J.~{Keppler}.
\newblock The impact of swebok version 3 on software engineering education and
  training.
\newblock In {\em 2014 IEEE 27th Conference on Software Engineering Education
  and Training (CSEE T)}, pages 192--200, 2014.

\bibitem{BOLZHINOSKI2019}
Darko Bozhinoski, Davide {Di Ruscio}, Ivano Malavolta, Patrizio Pelliccione,
  and Ivica Crnkovic.
\newblock {Safety for mobile robotic system: A systematic mapping study from a
  software engineering perspective}.
\newblock {\em Journal of Systems and Software}, 151:150--179, may 2019.

\end{thebibliography}

\end{document}